\documentclass[journal]{IEEEtran}
\usepackage{bbm}            
\usepackage{amsmath}
\usepackage{graphicx}
\graphicspath{ {./img/} }
\usepackage{multirow}
\usepackage{multicol}
\usepackage{xcolor}
\usepackage{tikz}
\usepackage{amssymb}
\usetikzlibrary{shapes,arrows,chains}

\begin{document}
%
\title{Improving Non-Intrusive Load Disaggregation through an Attention-Based Deep Neural Network}
%
%
%

\author{V. Piccialli and Antonio M. Sudoso\\
Department of Civil and Computer Engineering, University of Rome Tor Vergata, Rome, Italy}

\maketitle

\begin{abstract}
Energy disaggregation, known in the literature as Non-Intrusive Load Monitoring (NILM), is the task of inferring the power demand of the individual appliances given the aggregate power demand recorded by a single smart meter which monitors multiple appliances. In this paper, we propose a deep neural network that combines a regression subnetwork with a classification subnetwork for solving the NILM problem. Specifically, we improve the generalization capability of the overall architecture by including an encoder--decoder with a tailored attention mechanism in the regression subnetwork. The attention mechanism is inspired by the temporal attention that has been successfully applied in neural machine translation, text summarization, and speech recognition. The experiments conducted on two publicly available datasets---REDD and UK-DALE---show that our proposed deep neural network outperforms the state-of-the-art in all the considered experimental conditions. We also show that modeling attention translates into the network's ability to correctly detect the turning on or off an appliance and to locate signal sections with high power consumption, which are of extreme interest in the field of energy disaggregation.
\end{abstract}

\begin{IEEEkeywords}
Attention Mechanism, Deep Neural Network, Energy Disaggregation, Non-Intrusive Load Monitoring.
\end{IEEEkeywords}

%
\IEEEpeerreviewmaketitle

\section{Introduction}
%
%
%
%
Non-Intrusive Load Monitoring (NILM) is the task of estimating the power demand of each appliance given aggregate power demand signal recorded by a single electric meter monitoring multiple appliances~\cite{hart}. In~the last years, machine learning and optimization played a significant role in the research on NILM~\cite{ruano2019nilm, de2019load}. In~the literature, solutions based on k-Nearest Neighbor (k-NN),  Support Vector Machine (SVM), Matrix Factorization have been proposed~\cite{faustine2017survey, matrixFact}.
{A practical approach to NILM has to handle real power measurements sampled at intervals of seconds or minutes. In~this setting, one of the most popular approaches is based on the Hidden Markov Model (HMM) \cite{kolter2012approximate},
because of its ability to model transitions in consumption levels of real energy consumption for target appliances. Some successive papers focused on enhancing the  expressive power of this class of methods~\cite{makonin2015exploiting,nashrullah2019performance}. Recently, the~energy disaggregation problem has been reformulated as a multi-label classification problem~\cite{basu2014nonintrusive}. In~order to detect the active appliances at each time step, the~idea is to associate each value of the main power to a vector of labels of length equal to the number of appliances, that are set to 1 if the appliance is active and 0 otherwise. The~reformulated problem has been solved with different approaches~\cite{singhal2018simultaneous, singh2019non, multilabel:2020}. However, there is no direct way to derive the power consumption for each appliance at that time step.}
During the last years, approaches based on deep learning have received particular attention as they exhibited noteworthy disaggregation performance. Deep Neural Networks (DNNs) have been successfully applied for the first time to NILM by Kelly and Knottenbelt in~\cite{kelly2015neural}, who coined the term ``Neural NILM''. Neural NILM is a nonlinear regression problem that consists of training a neural network for each appliance in order to predict a  time window of the appliance load given the corresponding time window of aggregated data. Kelly and Knottenbelt proposed three different neural networks to perform NILM with high-frequency time series data: a recurrent neural network (RNN) using Long Short-Term Memory units (LSTM); a~Denoising Autoencoder (DAE); and~a regression model that predicts the start time, end time, and average power demand of each appliance. The~capability of LSTMs to successfully learn long-range temporal dependencies of time series data makes it a suitable candidate for NILM. Their first approach is based on stacked layers of LSTM units combined with a Convolutional Neural Network (CNN) at the beginning of the network to automatically extract features from the raw data. In~the same paper, NILM is treated as a noise reduction problem, in~which the disaggregated load represents the clean signal, and~the aggregated signal is considered corrupted by the presence of the remaining appliances and by the measurement noise. For~this purpose, noise reduction is performed by means of a DAE composed of convolutional layers and fully connected layers. In~the experiments conducted by the authors, the~DAE network outperforms the LSTM-based architecture and the other approaches frequently employed for this problem, such as HMMs and Combinatorial Optimization.
In~\cite{he2016empirical}, an~empirical investigation of deep learning methods is conducted by using two types of neural network architectures for NILM. The~first neural network solves a regression problem which estimates the transient power demand of a single appliance given the whole series of the aggregate power. The~second type of network is a multi-layer RNN using LSTM units, which is similar to the structure used in~\cite{kelly2015neural}.
Zhang~et~al.~\cite{zhang2018sequence} proposed instead a sequence-to-point learning for energy disaggregation where the midpoint of an appliance window is treated as classification output of a neural network with the aggregate window being the input.
Bonfigli~et~al.~\cite{bonfigli2018denoising} proposed different algorithmic and architecture improvements to the DAE for NILM, showing that the Neural NILM approach improves on the best known NILM approaches not based on DNNs like Additive Factorial Approximate Maximum A Posteriori estimation (AFAMAP) by Kolter and Jaakkola~\cite{kolter2012approximate}. Compared to~the work in \cite{kelly2015neural}, their DAE approach is improved by introducing pooling and upsampling layers in the architecture and a median filter in the disaggregation phase to reconstruct the output signal from the overlapped portions of the disaggregated signal. 
Shin~et~al.~\cite{shin2019subtask} proposed a subtask gated network (SGN) that combines two CNNs, namely, a regression subnetwork and a classification subnetwork. The~building block of the two subnetworks is the sequence-to-sequence CNN proposed in~\cite{zhang2018sequence}. In~their work, the~regression subnetwork is used to infer the initial power consumption, whereas the classification subnetwork focuses on the binary classification of the appliance state (on/off). The~final estimate of the power consumption is obtained by multiplying the regression output with the probability classification output. In~the experiments conducted by the authors, the~SGN outperforms HHMs and state-of-the-art CNN architectures that have been proposed recently~\cite{kelly2015neural, zhang2018sequence}.
Chen~et~al.~\cite{chen2019scale} adopted the structure of the SGN proposed in~\cite{shin2019subtask} and added to the SGN backbone a Generative Adversarial Network (GAN). In~their model, the~disaggregator for a given appliance is followed by a generator that produces the load pattern for that appliance. They show that adding the adversarial loss can help the model to produce more accurate result with respect to the basic SGN architecture.
None of these state-of-the art deep learning models use RNNs. In~fact, in~the NILM literature, CNNs have always shown better performance than RNNs, even though RNNs are still widely employed for sequence modeling tasks.  {In~\cite{postproc}, a~CNN-based DNN has been combined with data augmentation and an effective postprocessing phase, improving its ability to correctly detect the activation of each appliance with a small amount of data available.}
The attention mechanism applied to NILM is a relatively new idea~\cite{wang-attention}. The~DNN in~\cite{wang-attention} remarkably improves over Kelly's DAE when trained and tested on the same house. On~the other hand, the~generalization capability on houses not seen during the training is modest. Moreover, training and testing for the NILM task are time-consuming as they used the same architecture proposed in~\cite{bahdanau2014neural} for machine translation which consists of RNN layers in both the encoder and the~decoder.

In this paper, we propose a RNN-based encoder--decoder model to extract appliance specific power usage from the aggregated signal and we enhance it with a scalable and lightweight attention mechanism designed for the energy disaggregation task. More in detail, we substantially improve the generalization capability of the SGN by Shin~et~al. by encapsulating our model in the regression subnetwork and by combining it with the classification subnetwork.
The implemented attention mechanism has the function to strengthen the representational power of the neural network to locate the positions of the input sequence where the relevant information is present. The~intuition is that our attention-based model could help the energy disaggregation task by assigning importance to each position of the aggregated signal which corresponds to the position of a state change of the target appliance. This feature is crucial for developing appliance models that generalize well on buildings not seen during the~training.

The proposed DNN is tested on two publicly available datasets---REDD and UK-DALE---and~the performance is evaluated using different metrics. The~obtained results show that our algorithm outperforms state-of-the-art DNNs in all the addressed experimental conditions.
The paper is organized as follows. Section~\ref{sec:2} describes the NILM problem. Section~\ref{sec:3} presents our DNN architecture. Section~\ref{sec:4} describes the experimental procedure and the obtained results. Finally, Section~\ref{sec:5} concludes the~paper.

\section{NILM~Problem} \label{sec:2}
Given the aggregate power consumption $(x_1, \dots, x_T)$
from $N$ active appliances at the entry point of the meter, the~task of the NILM algorithm is to deduce the contribution $(y_1^i, \dots, y_T^i)$ of appliance $i = 1, \dots, N$, such that at time $t = 1, \dots, T$, the~aggregate power consumption is given by the sum of the power consumption of all the known appliances plus a noise term. The~energy disaggregation problem can be stated as
\begin{equation}
    x_t = \sum_{i=1}^{N} y_t^i + \epsilon_t,
\end{equation}
where $x_t$ is the aggregated active power measured at time $t$, $y_t^i$ is the individual contribution of appliance $i$, $N$ is the number of appliances, and~$\epsilon_t$ is a noise term. In~a denoised scenario, there is no noise term, whereas in a noised scenario, $\epsilon_t$ is given by the total contribution from appliances not included and the measurement noise. Similarly to~the work in \cite{kelly2015neural}, we refer to the power over a complete cycle of an appliance as an appliance activation. 
\section{Encoder--Decoder with Attention~Mechanism}  \label{sec:3}
In this section, we describe the adopted attention mechanism and DNN architecture for solving the NILM~problem. 

\subsection{Attention~Mechanism}
\begin{figure}[ht]
   \centering
   \includegraphics[scale=0.50]{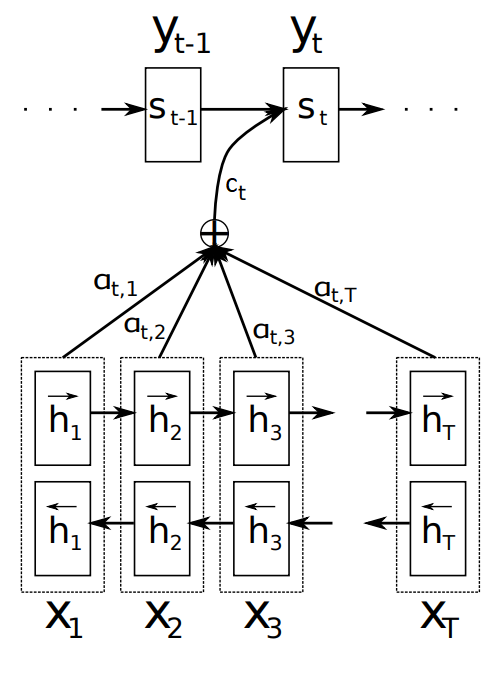}
   \caption{Original graphical representation of the attention model by Bahdanau~et~al. in~\cite{bahdanau2014neural}.}
   \label{fig:original_attention}
\end{figure}

In the classical setting, a~sequence-to-sequence network is a model consisting of two components called the encoder and decoder~\cite{sutskever2014sequence}. The~encoder is an RNN that takes an input sequence of vectors $(x_1, \dots, x_T)$, where $T$ is the length of input sequence, and~encodes the information into fixed length vectors $(h_1, \dots, h_T)$. This representation is expected to be a good summary of the entire input sequence. The~decoder is also an RNN which is initialized with a single context vector $c=h_T$ as its inputs and generates an output sequence $(y_1, \dots, y_{N})$ vector by vector, where $N$ is the length of output sequence. At~each time step, $h_t$ and $\sigma_t$ denote the hidden states of the encoder and decoder, respectively. There are two well-known challenges with this traditional encoder--decoder framework. First, a~critical disadvantage of single context vector design is the incapability of the system to remember long sequences: all the intermediate states of the encoder are eliminated and only the final hidden state vector is used to initialize the decoder. This technique works only for small sequences, however, as~the length of the sequence increases, the~vector becomes a bottleneck and may lead to loss of information~\cite{cho2014properties}. Second, it is unable to capture the need of alignment between input and output sequences, which is an essential aspect of structured output tasks such as machine translation or text summarization~\cite{young2018recent}.  
The attention mechanism, first introduced for machine translation by Bahdanau~et~al.~\cite{bahdanau2014neural}, was born to address these problems. The~novelty in their approach is the introduction of an alignment function that finds for each output word significant input words. In~this way, the~neural network learns to align and translate at the same time. The~central idea behind the attention is not to discard the intermediate encoder states but to combine and utilize all the states in order to construct the context vectors required by the decoder to generate the output sequence. This mechanism induces attention weights over the input sequence to prioritize the set of positions where relevant information is present. Following the definition from Bahdanau~et~al., attention-based models compute a context vector $c_t$ for each time step as the weighted sum of all hidden states of the encoder. Their corresponding attention weights are calculated as 
\begin{equation}
\footnotesize
    e_{tj} = f(\sigma_{t-1}, h_j), \quad \alpha_{tj} = \frac{exp(e_{tj})}{\sum_{k=1}^{T}exp(e_{tk})}, \quad c_t = \sum_{j=1}^{T} \alpha_{tj} h_j,
\end{equation}
where $f$ is a learned function that computes a scalar importance value for $h_j$ given the value of $h_j$ and the previous state $\sigma_{t-1}$ and each attention weight $\alpha_{tj}$ determines the normalized importance score for $h_j$. As~shown in Figure~\ref{fig:original_attention}, the~context vectors $c_t$ are then used to compute the decoder hidden state sequence, where $\sigma_t$ depends on $\sigma_{t-1}$, $c_t$, and $y_{t-1}$. The~attention weights can be learned by incorporating an additional feed-forward neural network that is jointly trained with encoder--decoder components of the~architecture.

The intuition is that an attention-based model could help in the energy disaggregation task by assigning importance to each position of the aggregated signal which corresponds to the position of an activation, or~more generally, to~a state change of the target appliance. This allows the neural network to focus its representational power on selected time steps of the target appliance in the aggregated signal, rather than on the activations of non-target appliances, hopefully yielding more accurate predictions. In~fact, some events (e.g., turning on or off an appliance) or signal sections (e.g., high power consumption) are more important than other parts within the input signal. For~this reason, being able to correctly detect the corresponding time steps can play a key role in the disaggregation task.
In neural machine translation, languages are typically not aligned because of the word ordering between the source and the target language. For~the NILM problem, the~aggregated power consumption is perfectly aligned with the load of the corresponding appliance and the alignment is known ahead of time. For~this reason, to~amplify the contribution of an appliance activation in the aggregated signal, we use a simplified attention model inspired by Raffel and Ellis~\cite{raffel2015feed}, that combines all the hidden states of the encoder using their relative importance. The~attention mechanism can be formulated as
\begin{equation}
    e_t = a(h_t), \quad \alpha_t = \frac{exp(e_t)}{\sum_{j=1}^{T}exp(e_j)}, \quad c = \sum_{t=1}^{T} \alpha_t h_t,
\end{equation}
where $a$ is a learnable function the depends only on the hidden state vector of the encoder $h_t$. The~function $a$ can be implemented with a feed-forward network that learns a particular attention weight $\alpha_{t}$ that determines the normalized importance score for $h_j$. This allows the network to recognize  the time steps that are more important to the desired output as the ones having higher attention~value.

\subsection{Model~Design}
From a practical point of view, DNNs use partial sequences obtained with a sliding window technique. The~duration of an appliance activation is used to determine the size of the window that selects the input and output sequences for the NILM modeling. To~be precise, let $\boldsymbol{x_{t, L}} = (x_t, \dots, x_{t+L-1})$ and $\boldsymbol{y_{t, L}^i} = (y_t^{i}, \dots y_{t+L-1}^i)$ be, respectively, the partial aggregate and appliance sequences of length $L$ starting at time $t$. In~addition, we build the auxiliary state sequence $(s_1^i, \dots, s_T^i)$, where $s_t^i \in \{0, 1\}$ represent the on/off state of the appliance $i$ at time $t$. The~state of an appliance is considered ``on'' when the consumption is greater than some threshold and ``off'' when the consumption is less or equal the same threshold. We use the notation $\boldsymbol{s_{t, L}^i} = (s_t, \dots, s_{t+L-1})$ for the partial state sequences of length $L$ starting at time $t$.
Our idea is to exploit the structure of the SGN architecture proposed in~\cite{shin2019subtask} as the building block of the model. This general framework uses an auxiliary sequence-to-sequence classification subnetwork that is jointly trained with a standard sequence-to-sequence regression subnetwork. The~difference here is that we generate a more accurate estimate of the power consumption by performing the regression subtask with a scalable RNN-based encoder--decoder with attention mechanism. The~intuition behind the proposed model is that the tailored attention mechanism allows the regression subnetwork to implicitly detect and assign more importance to some events (e.g., turning on or off of the appliance) and to specific signal sections (e.g., high power consumption), whereas the classification subnetwork helps the disaggregation process by enforcing explicitly the on/off~states.

Differently from the DNN in~\cite{wang-attention}, the~scalability of the overall architecture is ensured by the regression subnetwork where no RNN is needed in the decoder. In~fact, the~adopted attention mechanism allows one to decouple the input representation from the output and the structure of the encoder from the structure of the decoder. We exploit these benefits and we design a hybrid encoder--decoder which is based on a combination of convolutional layers and recurrent layers for the encoder and fully connected layers for the~decoder.

\subsection{Network~Topology}
 
Let $f_{reg}^i \colon \mathbb{R}_{+}^L \to \mathbb{R}_{+}^L$ be the appliance power estimation model, then the regression subnetwork learns the mapping $\boldsymbol{\hat{p}_{t, L}^i} = f_{reg}^i(\boldsymbol{x_{t, L}})$. The~topology of the regression subnetwork is as follows.

\textbf{Encoder}: The encoder network is composed by a CNN with 4 one-dimensional convolutional layers (Conv1D) with ReLU activation function that processes the input aggregated signal and extracts the appliance-specific signature as a set of feature maps. Finally, a RNN takes as input the set of feature maps and produces the sequence of the hidden states summarizing all the information of the aggregated signal. We use Bidirectional LSTM (BiLSTM) in order to get the hidden states $h_t$ that summarize the information from both directions. A~bidirectional LSTM is made up of a forward LSTM $\overrightarrow{g}$ that reads the sequence from left to right and a backward LSTM $\overleftarrow{g}$ that reads it from right to left. The~final sequence of the hidden states of the encoder is obtained by concatenating the hidden state vectors from both directions, i.e.,~$h_t=[\overrightarrow{h_t};\overleftarrow{h_t}]^T$. 

\begin{figure}[ht]
   \centering
   \includegraphics[scale=0.60]{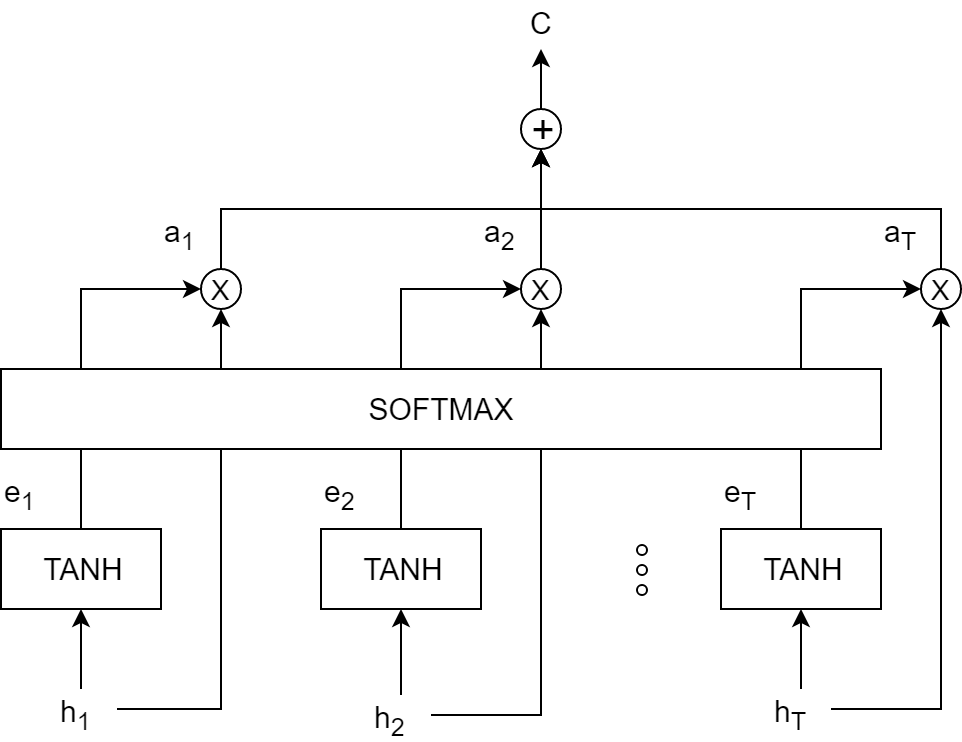}
   \caption{Graphical illustration of the implemented attention~unit.}
   \label{fig:attention_unit}
\end{figure}

\textbf{Attention}: The attention unit between the encoder and the decoder consists of a single layer feed-forward neural network that computes the attention weights and returns the context vector as a weighted average of the output of the encoder over time. Not all the feature maps produced by the CNN have equal contribution in the identification of the activation of the target appliance. Thus, the attention mechanism captures salient activations of the appliance, extracting more valuable feature maps than others for the disaggregation. 
{The implemented attention unit is shown in Figure~\ref{fig:attention_unit}, and it is mathematically defined as}
\begin{equation}
    e_t = V_a^T tanh(W_a h_t + b_a),
\end{equation}
\begin{equation}
    \alpha_t = softmax(e_t), 
\end{equation}
\begin{equation}
    c = \sum_{t=1}^{T} \alpha_t h_t,
\end{equation}
where $V_a$, $W_a$, and $b_a$ are the attentions parameters jointly learned with the other components of the architecture. The~output of the attention unit is the context vector $c$ that is used as the input vector for the following~decoder. 

\textbf{Decoder}: The decoder network is composed by 2 fully connected layers (Dense). The~second layer has the same number of units of the sequence length $L$.

The exact configuration of regression subnetwork is as~follows:

\begin{enumerate}
    \item Input (sequence length $L$ determined by the appliance duration)
    \item Conv1D (convolutional layer with $F$ filters, kernel size $K$, stride $1$, and~ReLU activation function)
    \item Conv1D (convolutional layer with $F$ filters, kernel size $K$, stride $1$, and~ReLU activation function)
    \item Conv1D (convolutional layer with $F$ filters, kernel size $K$, stride $1$, and~ReLU activation function)
    \item Conv1D (convolutional layer with $F$ filters, kernel size $K$, stride $1$, and~ReLU activation function)
    \item BiLSTM (bidirectional LSTM with $H$ units, and~tangent hyperbolic activation function)
    \item Attention (single layer feed-forward neural network with $H$ units, and~tangent hyperbolic activation function)
    \item Dense (fully connected layer with $H$ units, and~ReLU activation function)
    \item Dense (fully connected layer with $L$ units, and~linear activation function)
    \item Output (sequence length $L$)
\end{enumerate}

Let $f_{reg}^i \colon \mathbb{R}_{+}^L \to [0, 1]^L$ be the appliance state estimation model, then the
classification subnetwork learns the mapping $\boldsymbol{\hat{s}_{t, L}^i} = f_{cls}^i(\boldsymbol{x_{t, L}})$. We use the sequence-to-sequence CNN proposed in~\cite{zhang2018sequence} consisting of 6 convolutional layers followed by 2 fully connected layers. The~exact configuration of the classification subnetwork is the following:

\begin{enumerate}
    \item Input (sequence length $L$ determined by the appliance duration)
    \item Conv1D (convolutional layer with $30$ filters, kernel size $10$, stride $1$, and~ReLU activation function)
    \item Conv1D (convolutional layer with $30$ filters, kernel size $8$, stride $1$, and~ReLU activation function)
    \item Conv1D (convolutional layer with $40$ filters, kernel size $6$, stride $1$, and~ReLU activation function)
    \item Conv1D (convolutional layer with $50$ filters, kernel size $5$, stride $1$, and~ReLU activation function)
    \item Conv1D (convolutional layer with $50$ filters, kernel size $5$, stride $1$, and~ReLU activation function)
    \item Conv1D (convolutional layer with $50$ filters, kernel size $5$, stride $1$, and~ReLU activation function)
    \item Dense (fully connected layer with $1024$ units, and~ReLU activation function)
    \item Dense (fully connected layer with $L$ units, and~sigmoid activation function)
    \item Output (sequence length $L$)
\end{enumerate}

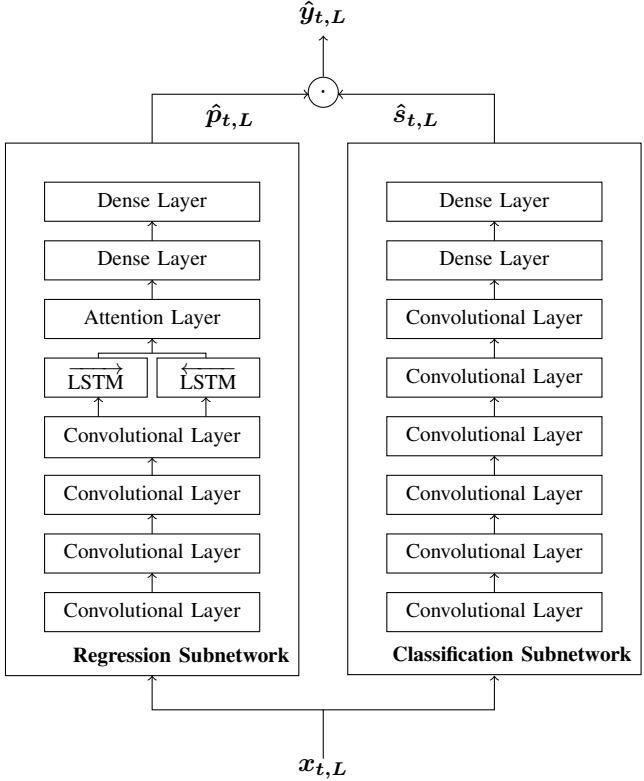
\begin{figure}[ht]
\begin{tikzpicture}[scale=1.30]
\draw (0,1) rectangle (3,-4.45)node[anchor = south east] {\footnotesize \textbf{Regression Subnetwork}};
\draw (0.4,0.2) rectangle (2.6,0.6)node[midway]{\footnotesize Dense Layer};
\draw[->](1.5,0) -- (1.5,0.2);
\draw (0.4,-0) rectangle (2.6,-0.4)node[midway]{\footnotesize Dense Layer};
\draw[<-](1.5,-0.4) -- (1.5,-0.6);
\draw (0.4,-0.6) rectangle (2.6,-1)node[midway]{\footnotesize Attention Layer};
\draw (0.95,-1.2) --(0.95,-1.15)--(2.05,-1.15) -- (2.05,-1.2);
\draw[<-] (1.5,-1) --(1.5,-1.15);
\draw (0.4,-1.6) rectangle (1.45,-1.2)node[midway]{\footnotesize $\overrightarrow{\textrm{LSTM}}$};
\draw[->](0.95,-1.8) --(0.95,-1.6);
\draw (1.55,-1.6) rectangle (2.6,-1.2)node[midway]{\footnotesize $\overleftarrow{\textrm{LSTM}}$};
\draw[->](2.05,-1.8) --(2.05,-1.6);
\draw (0.4,-1.8) rectangle (2.6,-2.22)node[midway]{\footnotesize Convolutional Layer};
\draw[->](1.5,-2.4) -- (1.5,-2.22);
\draw (0.4,-2.8) rectangle (2.6,-2.4)node[midway]{\footnotesize Convolutional Layer};
\draw[->](1.5,-3) -- (1.5,-2.8);
\draw (0.4,-3) rectangle (2.6,-3.4)node[midway]{\footnotesize Convolutional Layer};
\draw[<-](1.5,-3.4) -- (1.5,-3.6);
\draw (0.4,-3.6) rectangle (2.6,-4)node[midway]{\footnotesize Convolutional Layer};
\draw (3.5,1) rectangle (6.5,-4.44)node[anchor = south east] {\footnotesize \textcolor{white}{g}\textbf{Classification Subnetwork}};
\draw (3.9,0.60) rectangle (6.1,0.2)node[midway]{\footnotesize Dense Layer};
\draw[<-](5,0.2) -- (5,0);
\draw (3.9,-0.4) rectangle (6.1,0)node[midway]{\footnotesize Dense Layer};
\draw[->](5,-0.6) -- (5,-0.4);
\draw (3.9,-1) rectangle (6.1,-0.6)node[midway]{\footnotesize Convolutional Layer};
\draw[<-](5,-1) -- (5,-1.2);
\draw (3.9,-1.6) rectangle (6.1,-1.2)node[midway]{\footnotesize Convolutional Layer};
\draw (3.9,-1.8) rectangle (6.1,-2.20)node[midway]{\footnotesize Convolutional Layer};
\draw[->](5,-1.8) -- (5,-1.6);
\draw (3.9,-2.4) rectangle (6.1,-2.8)node[midway]{\footnotesize Convolutional Layer};
\draw[<-](5,-2.2) -- (5,-2.4);
\draw (3.9,-3) rectangle (6.1,-3.4)node[midway]{\footnotesize Convolutional Layer};
\draw[->](5,-3) -- (5,-2.8);
\draw (3.9,-3.6) rectangle (6.1,-4)node[midway]{\footnotesize Convolutional Layer};
\draw[<-](5,-3.4) -- (5,-3.6);
\draw (3.25,-5.4) node {$\boldsymbol{x_{t, L}}$};
\draw[->] (3.25,-5.3) -- (3.25,-4.8) -- (1.5,-4.8) -- (1.5,-4.45);
\draw[->]  (3.25,-4.8) -- (5,-4.8) -- (5,-4.45);
\draw (3.25,1.52) node[draw,circle,inner sep=0pt, minimum size=0.4cm]{$\cdot$};
\draw[->](1.5,1)--(1.5, 1.5)--node[anchor=north] {$\boldsymbol{\hat p_{t,L}}$}(3.10,1.5);
\draw[->](5,1)--(5, 1.5)--node[anchor=north] {$\boldsymbol{\hat s_{t,L}}$}((3.40,1.5);
\draw[->] (3.25,1.68)-- (3.25,2.1) node[anchor=south]{ $\boldsymbol{\hat y_{t,L}}$};
\end{tikzpicture}
\caption{Proposed Load Disaggregation with~Attention (LDwA) architecture used in our~experiments.}
\label{fig:network}
\end{figure}

The final estimate of  the  power  consumption  is obtained  by  multiplying  the regression  output  with  the  probability  classification  output:
\begin{equation}
    \boldsymbol{\hat{y}_{t. L}^i} = f_{out}^i(\boldsymbol{x_{t, L}}) = \boldsymbol{\hat{p}_{t, L}} \odot \boldsymbol{\hat{s}_{t, L}},
\end{equation}
where $\odot$ is the component-wise multiplication. The~overall architecture is shown in \mbox{Figure~\ref{fig:network}}, and~we call it LDwA, that is, Load Disaggregation with~Attention.

\unskip

\section{Experiments}  \label{sec:4}
In this section, we show the experiments performed to evaluate our LDwA approach. First, we describe the datasets, the~performance metrics, and the experimental procedure adopted. Then, we present and discuss the obtained~results. 

\subsection{Datasets}
In order to evaluate our algorithm and perform a fair comparison with state-of-the-art methods, we choose two publicly available real-world datasets and adopt the same partition into training and test sets of the previous studies~\cite{zhang2018sequence, shin2019subtask, chen2019scale}. The~Reference Energy Disaggregation Data Set (REDD) \cite{kolter2011redd} contains data for six houses in the USA at 1 second sampling period for the aggregate power consumption, and~at 3 s for the appliance power consumption. {Following the previous studies, we consider the 3 top-consuming appliances}: dishwasher (DW), fridge (FR), and microwave (MW). We use the data of houses 2--6 to build the training set, and~house 1 as the test set. The~preprocessed REDD dataset is provided by the authors of~\cite{shin2019subtask}.
The second dataset, the~Domestic Appliance-Level Electricity dataset UK-DALE~\cite{kelly2015uk}, contains over two years of consumption profiles of five houses in UK, at~a 6 s sampling period. {Here, the~experiments are conducted using the 5 top-consuming appliances}: dishwasher (DW), fridge (FR), kettle (KE), microwave (MW), and washing machine (WM). For~evaluation, we use houses 1, 3, 4, and 5 for training and house 2 for testing as in the previous works~\cite{zhang2018sequence, shin2019subtask, chen2019scale}. The~UK-DALE dataset has been preprocessed by the authors of~\cite{kelly2015neural}. We stress that for both datasets we consider the \textit{unseen} setting in which we train and test on different households. In~fact, the~best way to test the generalization capability of a model is to use the model on a building not seen during the training. This is a particularly desirable property for a NILM algorithm since the unseen scenario is more likely in the real world application of the NILM~service.

\subsection{Metrics}
In order to evaluate our NILM approach, we recall specific metrics that allow to capture significant performance of the algorithm. 
Following the previous studies in~\cite{zhang2018sequence, shin2019subtask, chen2019scale}, we use the Mean Absolute Error (MAE) and the Signal Aggregate Error (SAE). Let $y_i(t)$ and $\hat{y}_i(t)$ be the true power and the estimated power at time $t$ for the appliance $i$, respectively. The~MAE for the appliance $i$ is defined as
\begin{equation}
    \text{MAE}_i = \frac{1}{T} \sum_{t=1}^{T} |y_i(t) - \hat{y}_i(t)|.
\end{equation}

Give a predicted output sequence of length $T$, the~SAE for the appliance $i$ is defined~as
\begin{equation}
    \text{SAE}_i = \frac{1}{N} \sum_{\tau=1}^{N} \frac{1}{K} |r_i(\tau) - \hat{r}_i(\tau)|,
\end{equation}
where $N$ is the number of disjoint time periods of length $K$, $T = K \cdot N$, and $r_i(\tau)$ and $\hat{r}_i(\tau)$ represent the~sum of the true power and the sum of the predicted power in the $\tau$th time period, respectively
. In~our experiments, we set $N=1200$ which corresponds to a time period of approximately one hour for the REDD dataset and~two hours for the UK-DALE dataset. 
For both metrics, lower values indicate better disaggregation~performance.

{In order to measure how accurately each appliance is running in on/off states, we use classification metrics such as the F1-score, that is, the harmonic mean of precision (\emph{P}) and recall (\emph{R}):
\begin{equation}
    F_1 = \frac{ 2 P \cdot R}{P+R}, \quad P = \frac{TP}{TP + FP}, \quad R = \frac{TP}{TP + FN},
\end{equation}
where \emph{TP}, \emph{FP}, and \emph{FP} stand for true positive, false positive, and false negative, respectively. 
An appliance is considered ``on'' when the active power is greater than some threshold and ``off'' when it is less or equal the same threshold. The~threshold is assumed to be the same value used for extracting the activations~\cite{chen2019scale, shin2019subtask}. In~our  experiments, we use a threshold of 15 Watt for labeling the disaggregated loads. Precision, recall, and F1-score return a value between 0 and 1 where a higher number corresponds to better classification performance.}

\subsection{Network~Setup}
According to the Neural NILM approach, we train a network for each target appliance. A~mini-batch of 32 examples is fed to each neural network, and~mean and variance standardization is performed on the input sequences. For~the target data, min-max normalization is used where minimum and  maximum power consumption values of the related appliance are computed in the training set. The~training phase is performed with a sliding window technique over the aggregated signal, using overlapped windows of length $L$ with hop size equal to 1 sample. As~stated in~\cite{kelly2015neural}, the~window size for the input and output pairs has to be large enough to capture an entire appliance activation, but~not too large to include contributions of other appliances. In~Table~\ref{table:window_length}, we report the adopted window length $L$ for each appliance that is related to the dataset sampling rate. The~state classification label is generated by using a power consumption of 15 Watt as threshold. Each network is trained with the Stochastic Gradient Descent (SGD) algorithm with Nesterov momentum~\cite{sutskever2013importance} set to 0.9. The~loss function used for the joint optimization of the two subnetworks is given by $\mathcal{L} = \mathcal{L}_{out} + \mathcal{L}_{cls}$, where $\mathcal{L}_{out}$ is the Mean Squared Error (MSE) between the overall output of the network and the ground truth of a single appliance, and~$\mathcal{L}_{cls}$ is the binary cross-entropy (BCE) that measures the classification error of the on/off state for the classification subnetwork. The~maximum number of epochs is set to $100$, the~initial learning rate is set to $0.01$, and it is reduced by a decay factor equal to $10^{-6}$ as the training progresses. Early stopping is employed as a form of regularization to avoid overfitting since it stops the training when the error on the validation set starts to grow~\cite{early_stopping}. For~the classification subnetwork, we adopt the hyperparameters from~in \cite{shin2019subtask} as our focus is only the effectiveness of the proposed components. The~hyperparameter optimization of the regression subnetwork regards the number of filters ($F$), the~size of each kernel ($K$), and the number of neurons in the recurrent layer ($H$). Grid search is used to perform the hyperparameter optimization, which is simply an exhaustive search through a manually specified subset of points in the hyperparameter space of the neural network where $F=\{16, 32, 64\}$, $K=\{4, 8, 16\}$ and $H=\{256, 512, 1024\}$.
We evaluate the configuration of the hyperparameters on a held-out validation set and we choose the architecture achieving the highest performance on it. The~disaggregation phase, also carried out with a sliding window over the aggregated signal with hop size equal to 1 sample, generates overlapped windows of the disaggregated signal. Differently from what proposed in~\cite{kelly2015neural}, where the authors reconstruct the overlapped windows by aggregating their mean value, we adopt the strategy proposed in~\cite{bonfigli2018denoising} in which the disaggregated signal is reconstructed by means of a median filter on the overlapped portions. The~neural networks are implemented in Python with PyTorch, an~open source machine learning framework~\cite{pytorch} and the experiments are conducted on a cluster of NVIDIA Tesla K80 GPUs. The~training time requires several hours for each architecture depending on the network dimension and on the granularity of the~dataset.

\begin{table}[ht]
\centering
\begin{tabular}{|c|ccccc|}
\hline
\textbf{Dataset} & \textbf{DW} & \textbf{FR} & \textbf{KE} & \textbf{MW} & \textbf{WM} \\ \hline
REDD    & 2304 & 496 & - & 128 & - \\ \hline
UK-DALE & 1536 & 512 & 128 & 288 & 1024 \\ \hline
\end{tabular}
\caption{Sequence length ($L$) for the LDwA architecture.}
\label{table:window_length}
\end{table}

\subsection{Results}
We compare our approach with the HMM implemented in~\cite{batra2014nilmtk} and the DNNs recently proposed: DAE, Seq2Point, S2SwA, SGN, and SCANet. {We report the MAE, SAE, and F1-score for the REDD and the UK-DALE datasets in Tables~\ref{table:results_redd} and~\ref{table:results_ukdale}, respectively}. The~results show that our approach turns out to be by far the best for both datasets. Apart from us, the~two most competitive methods are SGN and SCANet, which share the same backbone we drew inspiration from. Results show that our network is better than both SGN and SCANet, implying that the differences introduced in our approach are significantly beneficial. {In particular, our network outperforms SGN, showing that including our regression network significantly improves both the estimate of the power consumption and the load classification, and~thus the overall disaggregation performance}. More in detail, for~the dataset REDD the improvement in terms of MAE (SAE) with respect to SGN ranges from a minimum of 24.13\% (23.44\%) on the fridge to a maximum of 45.15\% (54.4\%) on the dishwasher, with~an average improvement of 32.64\% (39.33\%). {As for the F1-score, the~classification performance increase from a minimum of 6.67\% on the fridge to a maximum of 24.03\% on the microwave, with~an average increase of 15.45\%.}
For the UK-DALE dataset instead, the~improvement in terms of MAE (SAE) with respect to SGN ranges from a minimum of 18.62\% (8.93\%) on the fridge to a maximum of 39.78\% (50.25\%) on the dishwasher, with~an average improvement of 27.84\% (30.65\%). {The F1-score increases from a minimum of 2.49\% on the kettle to a maximum of 10.82\% on the washing machine, with~an average increment of the accuracy of 6.79\%.} Moreover, our method outperforms the more recent SCANet getting better disaggregation performance on all the appliances for both the datasets and both the metric. Looking at the tables, we see that for the REDD dataset the improvement in terms of MAE (SAE) with respect to SCANet ranges from a minimum of 9\% (5.84\%) on the fridge to a maximum of 18.76\% (28.59\%) on the microwave, with~an average improvement of 13.21\% (15.03\%). {The improvement of the F1-score ranges from a minimum of 3.64\% on the fridge to maximum of 11.58\% on the microwave, with~an average increment of 6.81\% of the accuracy.} Finally, on~the UK-DALE dataset, the~improvement in terms of MAE (SAE) with respect to SCANet ranges from a minimum of 7.33\% (7.2\%) on the kettle to a maximum of 24.57\% (19.55\%) on the dishwasher, with~an average improvement of 15.69\% (14\%). {The F1-score increases from a minimum of 0.92\% on the kettle to a maximum of 8.85\% on the washing machine, with~an overall improvement of 4.41\%.}

\begin{figure}[!ht]
   \centering
   \includegraphics[scale=0.4]{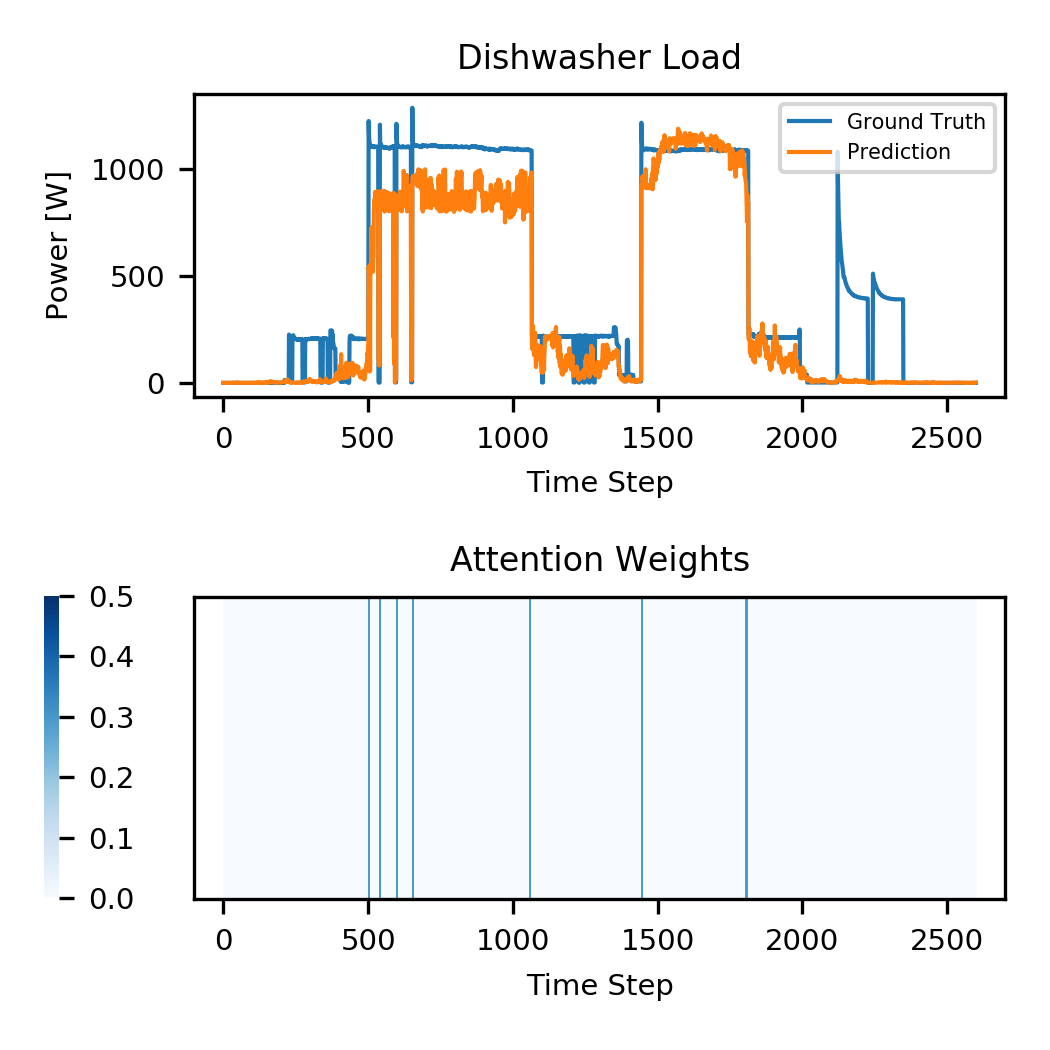}
   \caption{REDD dishwasher load and the heatmap of the attention weights at 3s resolution.}
   \label{fig:redd_dishwasher}
\end{figure}

\begin{figure}[!ht]
   \centering
   \includegraphics[scale=0.4]{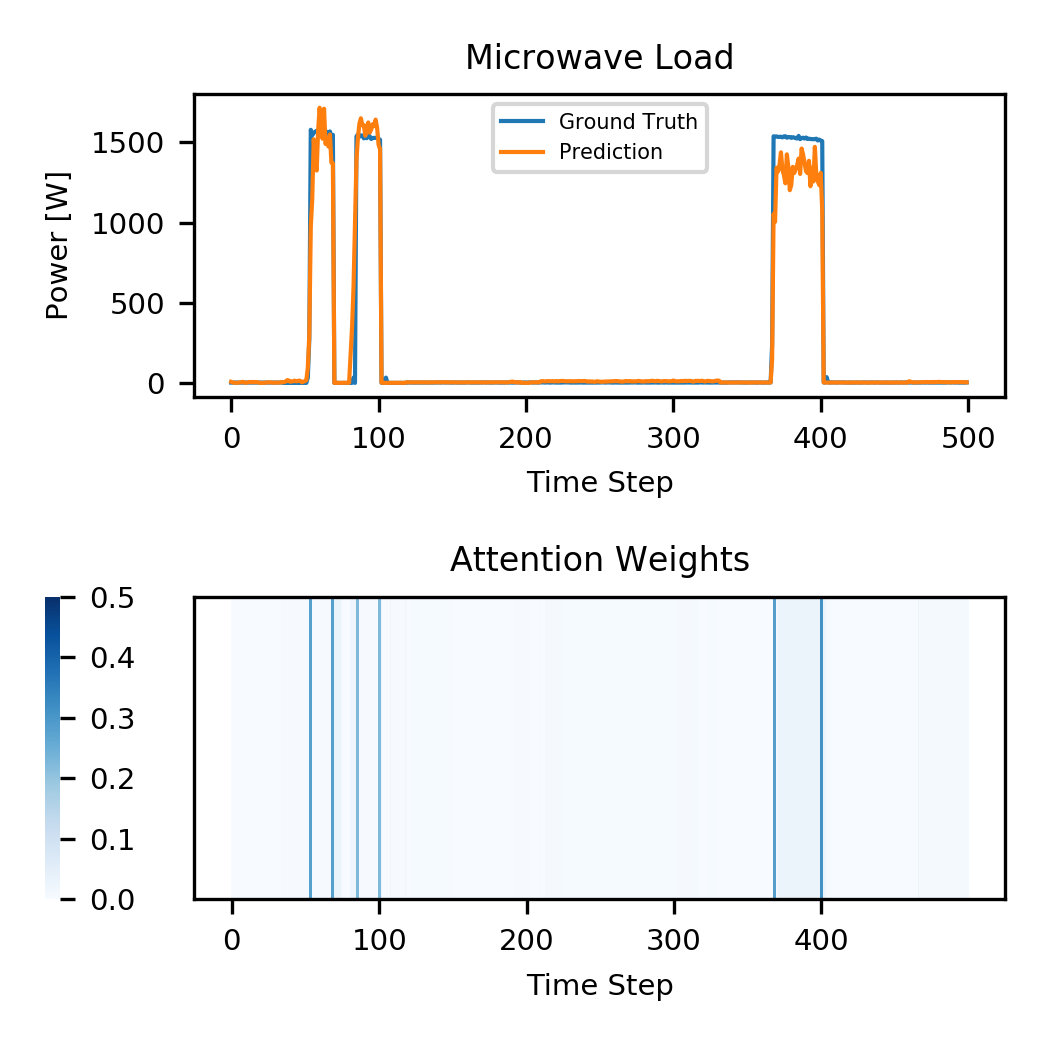}
   \caption{REDD microwave load and the heatmap of the attention weights {at 3s resolution}.}
   \label{fig:redd_microwave}
\end{figure}

\begin{figure}[!ht]
   \centering
   \includegraphics[scale=0.4]{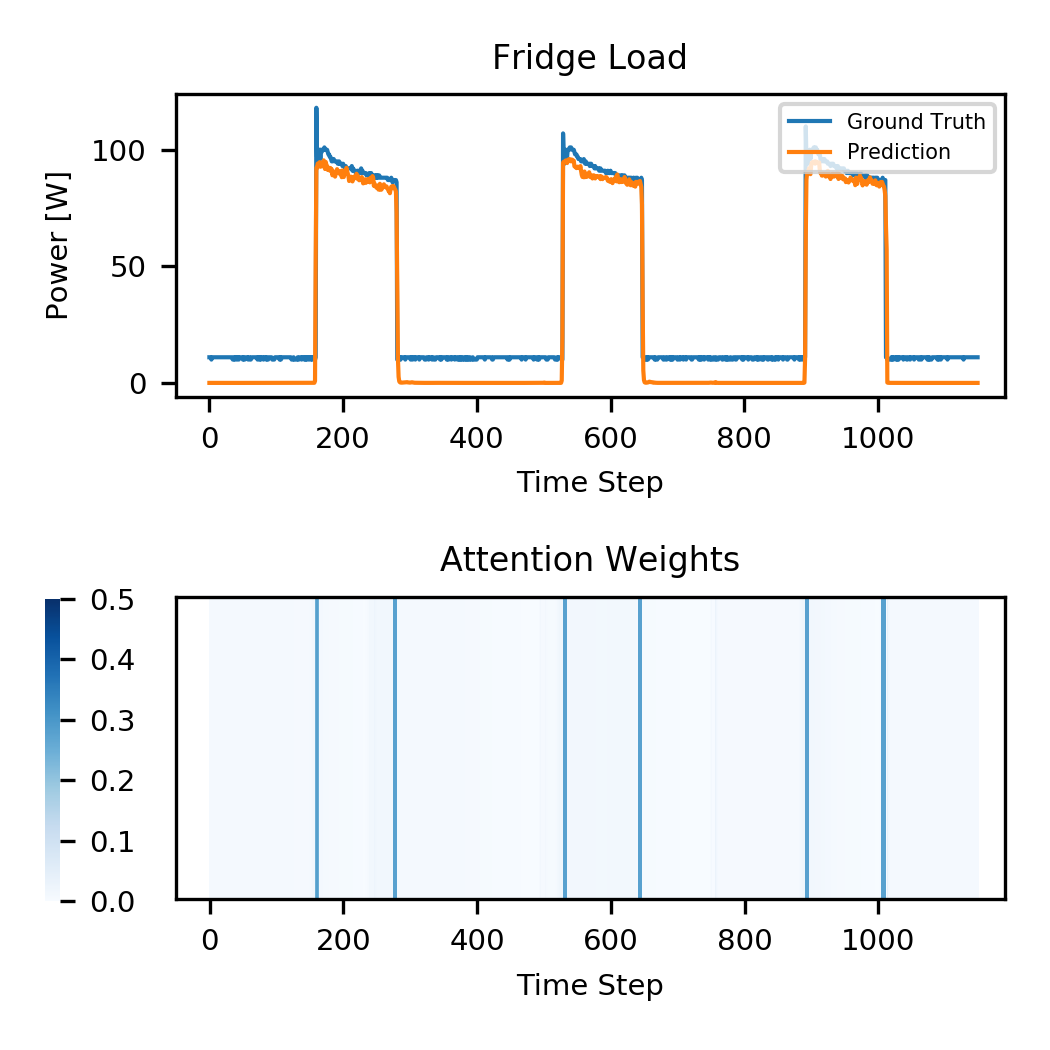}
   \caption{UK-DALE fridge load and the heatmap of the attention weights at 6s resolution.}
   \label{fig:ukdale_fridge}
\end{figure}

\begin{figure}[!ht]
   \centering
   \includegraphics[scale=0.4]{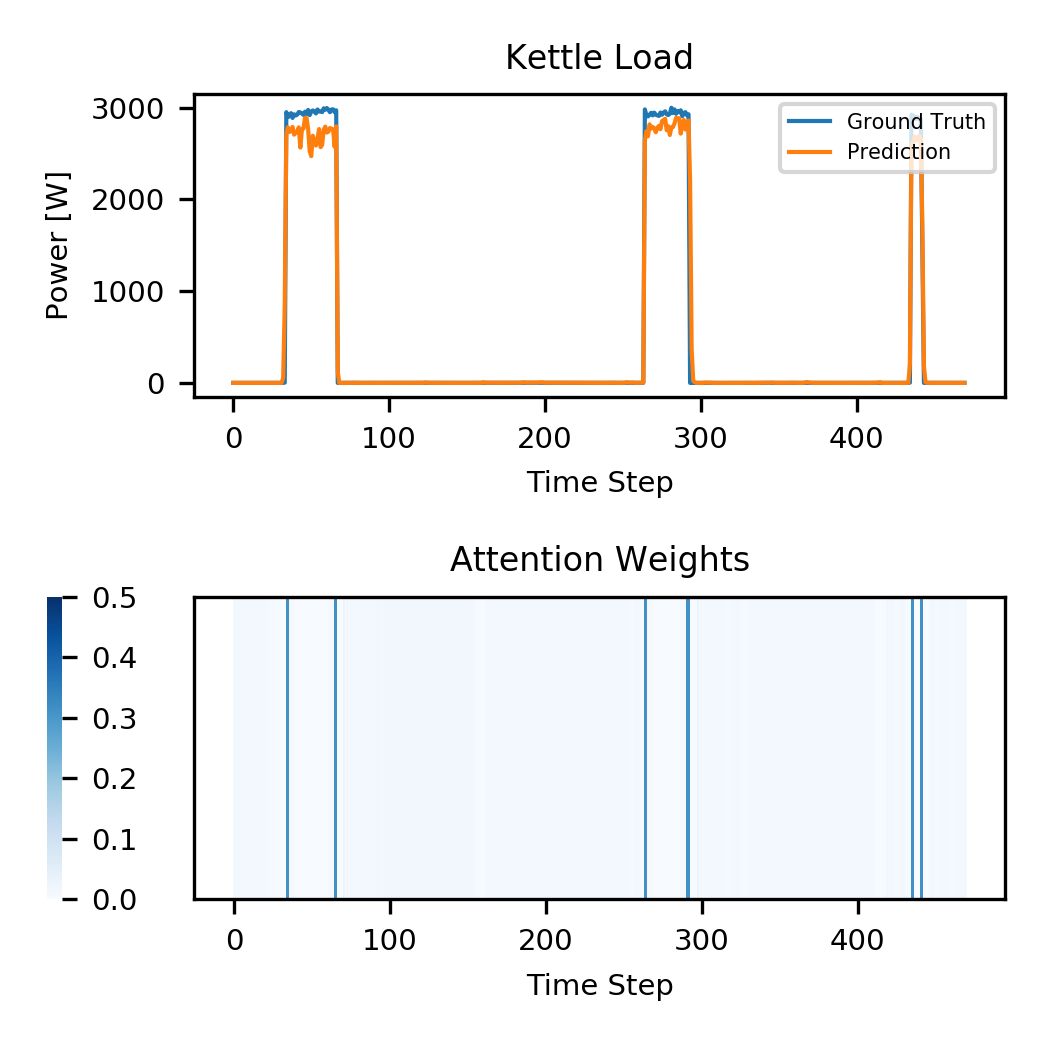}
   \caption{UK-DALE kettle load and the heatmap of the attention weights at 6s resolution.}
   \label{fig:ukdale_kettle}
\end{figure}

In order to evaluate the computational burden of the proposed LDwA, we report in Tables~\ref{tab:training_redd} and~\ref{tab:training_ukdale} the training time with respect to the most accurate DNNs. Clearly, LDwA is less efficient than SGN as LSTM layers have larger number of trainable parameters than the convolutional ones. However, the~efficiency of our architecture with respect to the attention-based S2SwA is remarkable. This is explained by the presence of the tailored attention mechanism that does not require additional recurrent layers in the decoder. There is also a huge improvement in the training time with respect to the SCANet. We achieve better performance without the need to train a Generative Adversarial Network, that requires a significant amount of computational resources and has notorious convergence~issues.
The profiles related to the dishwasher, microwave, fridge, and kettle are shown in Figures~\ref{fig:redd_dishwasher}--\ref{fig:ukdale_kettle}, respectively, where each appliance activation is successfully detected by the  LDwA in the disaggregated~trace. The tailored attention mechanism inserted into the regression branch of the network allows us to correctly identify the relevant time steps in the signal and generalize well in unseen houses. Furthermore, modeling attention is particularly interesting from the perspective of the interpretability of deep learning models because it allows one to directly inspect the internal working of the architecture. The~hypothesis is that the magnitude of the attention weights correlates with how relevant the specific region of the input sequence is, for~the prediction of the output sequence. As~shown in Figures~\ref{fig:redd_dishwasher}--\ref{fig:ukdale_kettle}, our network is effective at predicting the activation of an appliance and the attention weights present a peak in correspondence of the state change of that~appliance.
In conclusion, our approach does not only predict the correct disaggregation in terms of scale, but~is also successful at deciding if the target appliance is active in the aggregate load at a given time~step.

\begin{table}[!ht]
\begin{center}
\begin{tabular}{ |c|c|ccc|c| } 
\hline
\textbf{Model} & \textbf{Metric} & \textbf{DW}  & \textbf{FR} & \textbf{MW} & \textbf{Overall}\\
\hline
\multirow{3}{*}{FHMM \cite{batra2014nilmtk}}
 & MAE & 101.30 & 98.67 & 87.00 & 95.66 \\ 
 & SAE & 93.64  & 46.73 & 65.03 & 68.47 \\ 
 & {F1 (\%)} & {12.93} & {35.12} & {11.97} & {20.01} \\\hline
\multirow{3}{*}{DAE \cite{bonfigli2018denoising}} 
 & MAE & 26.18 & 29.11 & 23.26 & 26.18 \\  
 & SAE & 21.46 & 20.97 & 19.14 & 20.52 \\
 & {F1 (\%)} & {48.81} & {74.76} & {18.54} & {47.37} \\\hline
\multirow{3}{*}{Seq2Point \cite{zhang2018sequence}}
 & MAE & 24.44  & 26.01 & 27.13 & 25.86 \\ 
 & SAE & 22.87  & 16.24 & 18.89 & 19.33 \\
 & {F1 (\%)} & {47.66} & {75.12} & {17.43} & {46.74} \\\hline
 \multirow{3}{*}{S2SwA \cite{wang-attention}}
 & MAE & 23.48  & 25.98 & 24.27 & 24.57 \\ 
 & SAE & 22.64  & 17.26 & 16.19 & 18.69 \\
 & {F1 (\%)} & {49.32} & {76.98} & {19.31} & {48.57} \\\hline
\multirow{3}{*}{SGN \cite{shin2019subtask}} 
 & MAE & 15.77  & 26.11 & 16.95 & 19.61 \\ 
 & SAE & 15.22  & 17.28 & 12.49 & 15.00 \\
 & {F1 (\%)} & {58.78} & {80.09} & {44.98} & {61.28} \\\hline
\multirow{3}{*}{SCANet \cite{chen2019scale}} 
 & MAE & 10.14 & 21.77 & 13.75 & 15.22 \\  
 & SAE & 8.12 & 14.05 & 9.97 & 10.71 \\
 & {F1 (\%)} & {69.21} & {83.12} & {57.43} & {69.92} \\\hline
\multirow{3}{*}{Proposed LDwA} 
 & MAE & \textbf{8.65} & \textbf{19.81} & \textbf{11.17} & \textbf{13.21} \\ 
 & SAE & \textbf{6.94} & \textbf{13.23} & \textbf{7.12} & \textbf{9.10} \\
 & {F1 (\%)} & {\textbf{74.41}} & {\textbf{86.76}} & {\textbf{69.01}} & {\textbf{76.73}} \\\hline
\end{tabular}
\caption{Disaggregation performance for the REDD dataset. We report in boldface the best approach.}
\label{table:results_redd}
\end{center}
\end{table}

\begin{table}[!ht]
\begin{center}
\resizebox{\columnwidth}{!}{%
\begin{tabular}{ |c|c|ccccc|c| } 
\hline
\textbf{Model} & \textbf{Metric} & \textbf{DW}  & \textbf{FR} & \textbf{KE} & \textbf{MW} & \textbf{WM} & \textbf{Overall}\\
\hline
\multirow{3}{*}{FHMM \cite{batra2014nilmtk}}
 & MAE & 48.25 & 60.93 & 38.02 & 43.63 & 67.91 & 51.75\\ 
 & SAE & 46.04 & 51.90 & 35.41 & 41.52 & 64.15 & 47.80\\
 & {F1 (\%)} & {11.79} & {33.52} & {9.35} & {3.44} & {4.10} & {12.44} \\ \hline
\multirow{3}{*}{DAE \cite{bonfigli2018denoising}} 
 & MAE & 22.18 & 17.72 & 10.87 & 12.87 & 13.64 & 15.46\\ 
 & SAE & 18.24 & 8.74 & 7.95 & 9.99 & 10.67 & 11.12\\
 & {F1 (\%)} & {54.88} & {75.98} & {93.43} & {31.32} & {24.54} & {56.03} \\\hline
\multirow{3}{*}{Seq2Point \cite{zhang2018sequence}} 
 & MAE & 15.96 & 17.48 & 10.81 & 12.47 & 10.87 & 13.52\\ 
 & SAE & 10.65 & 8.01 & 5.30 & 10.33 & 8.69 & 8.60\\
 & {F1 (\%)} & {50.92} & {80.32} & {94.88} & {45.41} & {49.11} & {64.13}\\\hline
 \multirow{3}{*}{S2SwA \cite{wang-attention}} 
 & MAE & 14.96 & 16.47 & 12.02 & 10.37 & 9.87 & 12.74\\ 
 & SAE & 10.68 & 7.81 & 5.78 & 8.33 & 8.09 & 8.14\\
 & {F1 (\%)} & {53.67} & {79.04} & {94.62} & {47.99} & {45.79} & {64.22}\\\hline
\multirow{3}{*}{SGN \cite{shin2019subtask}} 
 & MAE & 10.91 & 16.27 & 8.09 & 5.62 & 9.74 & 10.13\\ 
 & SAE & 7.86 & 6.61 & 5.03 & 4.32 & 7.14 & 6.20\\
 & {F1 (\%)} & {60.02} & {84.43} & {96.32} & {58.55} & {61.12} & {72.09}\\\hline
\multirow{3}{*}{SCANet \cite{chen2019scale}} 
 & MAE & 8.71 & 15.16 & 6.14 & 4.82 & 8.48 & 8.67\\ 
 & SAE & 4.86 & 6.54 & 4.03 & 3.81 & 5.77 & 5.00\\
 & {F1 (\%)} & {63.30} & {85.77} & {98.89} & {62.22} & {63.09} & {74.65}\\\hline
\multirow{3}{*}{Proposed LDwA} 
 & MAE & \textbf{6.57} & \textbf{13.24} & \textbf{5.69} & \textbf{3.79} & \textbf{7.26} & \textbf{7.31}\\ 
 & SAE & \textbf{3.91} & \textbf{6.02} & \textbf{3.74} & \textbf{2.98} & \textbf{4.87} & \textbf{4.30}\\
 & {F1 (\%)} & {\textbf{68.99}} & {\textbf{87.01}} & {\textbf{99.81}} & {\textbf{67.55}} & {\textbf{71.94}} & {\textbf{79.06}}\\\hline
\end{tabular}}
\caption{Disaggregation performance for the UK-DALE dataset. We report in boldface the best approach.}
\label{table:results_ukdale}
\end{center}
\end{table}

\begin{table}[!ht]
    \centering
    \begin{tabular}{ |c|ccc| }
    \hline
    \textbf{Model} & \textbf{DW}  & \textbf{FR} & \textbf{MW} \\\hline
    S2SwA \cite{wang-attention} &  8.91 & 5.75 & 3.93\\
    SGN   \cite{shin2019subtask} &  3.31 & 2.43 & 0.57\\
    SCANet \cite{chen2019scale} &  6.44 & 3.24 & 2.68\\
    Proposed LDwA & 5.37 & 3.07 & 1.88\\\hline
    \end{tabular}
    \caption{{Training time in hours for the REDD dataset.}}
    \label{tab:training_redd}
\end{table}

\begin{table}[!ht]
    \centering
    \begin{tabular}{ |c|ccccc| }
    \hline
    \textbf{Model} & \textbf{DW}  & \textbf{FR} & \textbf{KE} & \textbf{MW} & \textbf{WM} \\\hline
    S2SwA  \cite{wang-attention} &  14.53 & 9.42 & 5.02 & 6.21 & 6.12 \\
    SGN   \cite{shin2019subtask} &  5.43 & 3.11 & 2.43 & 2.76 & 3.78 \\
    SCANet \cite{chen2019scale} & 8.01 & 7.98 & 6.41 & 5.77 & 7.11 \\
    Proposed LDwA & 6.21 & 4.93 & 3.65 & 3.37 & 5.87 \\\hline
    \end{tabular}
    \caption{{Training time in hours for the UK-DALE dataset.}}
    \label{tab:training_ukdale}
\end{table}

\section{Conclusions}  \label{sec:5}
This paper proposes LDwA, a~new deep neural network architecture for the NILM problem that features a tailored attention mechanism with the encoder--decoder framework to extract appliance specific power usage from the aggregated signal. The~integration of convolutional layers and recurrent layers in the regression subnetwork facilitates feature extraction and allows to build better appliance models where the locations of relevant features are successfully identified by the attention mechanism. The~use of the proposed model for the regression subtask increases the network's ability to extract and exploit information dramatically. The~proposed system is tested on two real-world datasets with different granularity, REDD and UK-DALE. The~experimental results demonstrate that the proposed model significantly improves accuracy and generalization capability for load recognition of all the appliances for both datasets compared to the deep learning~state-of-the-art.

\newpage


\bibliographystyle{IEEEtran}
%

\bibliography{mybibfile}

%

\end{document}